\documentclass[conference,a4paper]{IEEEtran}
\IEEEoverridecommandlockouts

\usepackage{multirow}
\usepackage{subcaption} 
\usepackage{adjustbox}
\usepackage{amsmath,amssymb,amsfonts}
\usepackage{fancyhdr}
\usepackage{algorithmic}
\usepackage{graphicx}
\usepackage{textcomp}
\usepackage{xcolor}
\usepackage{graphicx}
\usepackage{comment}
\usepackage{hyperref}
\usepackage{pifont}

\usepackage{amssymb}
\usepackage{mathrsfs}
\usepackage{calrsfs}

\usepackage[numbers,sort&compress]{natbib}

\def\BibTeX{{\rm B\kern-.05em{\sc i\kern-.025em b}\kern-.08em
    T\kern-.1667em\lower.7ex\hbox{E}\kern-.125emX}}

\usepackage{calrsfs}
\DeclareMathAlphabet{\pazocal}{OMS}{zplm}{m}{n}

\newcommand{\Lb}{\pazocal{L}}
\newcommand{\etal}{\textit{et al.}}
\newcommand{\cmmnt}[1]{}

\begin{document}

\title{
Locally-Focused Face Representation for Sketch-to-Image Generation Using Noise-Induced Refinement 
}

\author{
   \small
   Muhammad Umer Ramzan\textsuperscript{1}, Ali Zia\textsuperscript{2,3}, Abdelwahed Khamis\textsuperscript{3}, 
   Ayman Elgharabawy\textsuperscript{2}, Ahmad Liaqat\textsuperscript{1}, Usman Ali\textsuperscript{1} \\
   \textsuperscript{1}GIFT University, Gujranwala, Pakistan \textsuperscript{2}Australian National University, Australia 
   \textsuperscript{3}CSIRO, Australia \\
   \{umer.ramzan, usmanali, 191370062\}@gift.edu.pk, \{ali.zia, ayman.Gh\}@anu.edu.au, abdelwahed.khamis@data61.csiro.au
}

\maketitle

\newcommand{\leftfooter}{979-8-3503-7903-7/24/\$31.00 ©2024 IEEE}

\fancypagestyle{myfooter}{
  \fancyfoot[L]{\leftfooter} 
  \renewcommand{\headrulewidth}{0pt} 
  \renewcommand{\footrulewidth}{0pt} 
  }
\thispagestyle{myfooter}

\begin{abstract}
This paper presents a novel deep-learning framework that significantly enhances the transformation of rudimentary face sketches into high-fidelity colour images. Employing a Convolutional Block Attention-based Auto-encoder Network (CA2N), our approach effectively captures and enhances critical facial features through a block attention mechanism within an encoder-decoder architecture.  Subsequently, the framework utilises a noise-induced conditional Generative Adversarial Network (cGAN) process that allows the system to maintain high performance even on domains unseen during the training. These enhancements lead to considerable improvements in image realism and fidelity, with our model achieving superior performance metrics that outperform the best method by FID margin of 17, 23, and 38 on CelebAMask-HQ, CUHK, and CUFSF datasets; respectively. The model sets a new state-of-the-art in sketch-to-image generation, can generalize across sketch types, and offers a robust solution for applications such as criminal identification in law enforcement.
\end{abstract}


\section{\textbf{Introduction}}
Sketching serves as an intuitive method for rapidly depicting objects or scenes, with diverse applications across various fields. Notably, sketching plays a critical role in law enforcement, particularly in the identification and apprehension of criminals~\cite{More2023SketchBasedIR}. However, sketches produced by eyewitnesses or non-professional artists typically exhibit simplicity and imperfections. These sketches often lack crucial details and exhibit misalignment in stroke placement relative to the original images, leading to a sparsity of features. Addressing these limitations, our research introduces a novel approach using a Convolutional Block Attention-based Auto-encoder Network (CA2N) specifically designed to enhance \emph{locally-focused} face representation. This network independently identifies and processes five distinct facial feature descriptors: left eye, right eye, nose, mouth, and other facial regions, thereby refining the quality and accuracy of the initial sketches. Our goal is to bridge the gap between rudimentary sketches and high-fidelity image generation, ensuring that crucial facial components are accurately represented and reconstructed in the subsequent sketch-to-image generation process. Furthermore, the incorporation of new loss functions with \emph{noise-induced} refinement techniques helps in achieving enhanced detail and realism in the generated images, promoting better model generalization and robustness.

The transformation of basic sketches into accurate images has been significantly advanced by developments in image processing and machine learning, notably through the introduction of Generative Adversarial Networks (GANs) that enable precise manipulation of facial attributes via integrated encoder-decoder architectures~\cite{Wu2023SketchSceneSS}. Conditional Generative Adversarial Networks (cGANs) further enhance this process by embedding contextual information within traditional GAN architectures~\cite{Hu2020FacialAS}, while Deep Convolutional Generative Adversarial Networks (DCGAN) have improved training stability and image fidelity~\cite{mathew2021deep,chen2020deepfacedrawing}. Despite these advancements, challenges such as information loss and image blurriness persist, often due to the formation of manifolds within the sketch's embedding space that reflects similar geometries across domains~\cite{liu2005nonlinear}. Techniques to construct multichannel feature maps and sketch refinement strategies have been developed to enhance detail accuracy without relying on sketch-based training data~\cite{Yang2021ControllableST}. To mitigate significant detail obscuration and improve feature reproduction, the Sketch-Guided Latent Diffusion Model (SGLDM) uses a multi-autoencoder (MAE) to capture and replicate detailed facial features from degraded sketches~\cite{Peng2024SketchGuidedLD}, although it can still generate noise and artefacts in low-quality sketches.

Our framework adopts a two-stage learning process to address the challenges associated with the absence of critical image information and artifacts. The employment of a two-stage network is particularly effective in sketch-to-image tasks, as supported by literature indicating that such architectures allow for more nuanced handling of complex image characteristics and better adaptation to varying input data conditions~\cite{liu2020unsupervised}. Initially, feature representation learning is conducted through CA2N, tailored to enhance the selection and representation of features. This network meticulously identifies five distinct feature descriptors from the facial sketch data, corresponding to the left eye, right eye, nose, mouth, and the remaining facial regions. These descriptors are crucial for ensuring that each facial component is captured with high fidelity, setting the stage for more accurate subsequent image generation.

Following feature extraction, our approach transitions into the second stage, which involves a noise-inducted adaptation learning process. This phase utilises the feature vectors extracted earlier and transforms them into spatial feature maps. These maps are then fed into a cGAN, which facilitates the production of color images. The cGAN is designed to optimize image generation through the application of global and local losses, leveraging structural similarity index and L1 loss to refine the output. This method not only enhances the realism of the generated images but also ensures robustness against domain variability. The introduction of noise plays a pivotal role in this process, particularly in mitigating the typical blurriness associated with low-resolution images produced by baseline methods, and strengthens our model's capacity to maintain consistency across different input variations~\cite{osowiechi2024nc}.
To further enhance the visual quality and fine-grain details of the generated images, an image enhancement module utilising Generative Facial Prior Generative Adversarial Networks (GFPGANs)~\cite{wang2021towards} has been integrated into our architecture. This module significantly improves the clarity and detail of the images, making them more visually appealing and finely tuned. The main contributions of our proposed framework are outlined as follows:
\begin{enumerate}
\item \textbf{Feature Extraction:} Utilising a block attention approach within an encoder-decoder architecture to capture and enhance the relationship between localized facial features effectively.

\item \textbf{Image Realism and Fidelity:} Applying an iterative refinement technique using cGAN, which includes a noise-induction domain adaptation learning process sourced from a random uniform distribution, to improve the realism and fidelity of the generated images significantly.

\item \textbf{Innovative Loss Functions:} Introducing  a novel global sketch to image objective function that combines losses such as  the cGAN loss, the perceptual loss for structural similarity index and $\Lb_1$ loss; respectively. These are crucial in optimizing the training process.

\item \textbf{Model Generalization:} Ensuring the adaptability of the learning model across various sketch types, such as hand-drawn, line, and Photoshop sketches, facilitated by the strategic implementation of noise induction.
\end{enumerate}

Following the introduction, Section \ref{sec:related_work} reviews related work in sketch-to-image conversion, defining the context and importance of our contributions. Section \ref{sec:method} details the methodology, including the architecture of the Convolutional Block Attention-based Auto-encoder Network (CA2N) and the noise-induced domain adaptation learning process. Section \ref{sec:results and discussion} presents the experimental results through quantitative and qualitative analyses. Section \ref{sec:Conclusion} concludes by summarising the key contributions and discussing the impact of our work on the field of sketch-to-image synthesis.

\section{\textbf{Related Work}}
\label{sec:related_work}
The literature on sketch-to-image translation, particularly within the criminal record domain, has seen varied approaches leveraging advanced neural architectures. Notably, More \etal~\cite{More2023SketchBasedIR} explored the use of Deep Convolutional Neural Networks (DCNN) to facilitate the transformation from sketches to realistic images. Meanwhile, general translation challenges have been addressed through diverse methodologies aimed at increasing realism from input sketches~\cite{chen2018sketchygan}. A significant development in this area has been the pix2pix framework introduced by Isola \etal~\cite{isola2017image}, which employs a U-Net based cGAN. Despite its innovative approach, pix2pix often produces low-resolution images (256x256), which impacts the quality of the output. This limitation was addressed by Wang \etal~\cite{wang2018high} with Pix2PixHD, which enhances image quality by splitting the generator into two sub-networks trained on high-resolution images.

While these methods have demonstrated success in generating visually impressive images from sketches, they often fail to capture accurate structural details such as specific facial features, resulting in outputs that do not closely resemble the original sketches~\cite{isola2017image,wang2018high}. To counteract this, Li \etal~\cite{li2019linestofacephoto} introduced the LinesToFacePhoto framework, utilizing a conditional self-attention mechanism within a generative adversarial network to improve the realism of face images generated from line sketches. Further refinements in this domain have been introduced by Zhao \etal~\cite{zhao2019generating}, who developed a method to control specific facial attributes such as gender and age in the generated images. Meanwhile, Hicsonmez \etal~\cite{hicsonmez2023improving} proposed a technique that enhances the colorization accuracy of sketches, employing an adversarial segmentation loss to better align colorized images with sketch boundaries. This approach addresses the frequent misalignments between colorized outputs and their corresponding sketches, enhancing the visual realism of the results.

Additionally, novel methodologies continue to emerge, such as the SketchInverter by An \etal~\cite{an2023sketchinverter}, which employs a cGAN architecture to handle multi-class sketches effectively, improving upon existing techniques by using a conditional encoder that maps sketches into a latent space. This approach ensures that the generated images maintain the structural integrity of the original sketches. Moreover, Richardson \etal~\cite{richardson2021encoding} introduced pixel2style2pixel (pSp), an image translation framework that directly maps images into a latent space, which is then used to generate stylized outputs without requiring iterative optimization. This method, however, is constrained by the types of images StyleGAN can generate, limiting its flexibility.

Recent advancements by Guo \etal~\cite{guo2024image} with the Offset-based Multi-Scale Codes GAN Encoder (OME) demonstrate significant improvements in image-to-image translation. OME integrates a multi-scale latent codes GAN encoder with an iterative update synthesis approach, enhancing the generation of detailed and realistic images. However, the dependency on pre-trained GANs restricts OME’s ability to accurately render certain features, which could limit its application in precise sketch-based image synthesis tasks.

Building upon the existing research, our work specifically addresses the persistent challenge of achieving high fidelity and structural accuracy in sketch-to-image translation, a gap highlighted by prior studies. Our approach utilizes a two-stage learning process with a Convolutional Block Attention-based Auto-encoder Network (CA2N) and a noise-induced domain adaptation strategy, details are given in the next section.

\section{\textbf{Two Stage Sketch to Image Face Generation}}
\label{sec:method}
Our sketch-to-image method (Fig.~\ref{fig:architecture}) comprises two stages encompassing a representation learning followed by adversarial generation. In the first stage, we learn a powerful representation of individual facial sketch components that can utilized in the subsequent stage for image generation. Then, in the second stage, adversarial generation conditioned on the learned facial embeddings  (from stage 1) is used to produce the final image. Notably, we augment conditional GAN with various losses including a noise-induced loss to improve the generalization.

There are two unique aspects of our design. First,  we opt for a locally-focused representation of facial sketches. This stands in contrast to whole-sketch to whole-image transformation approaches \cite{wang2018high}, \cite{zhu2017unpaired}. It allows obtaining faithful representation that preserves the details of various facial components. Yet, it can results in disconnected poor embeddings as the components are individually processed. An issue for which we integrate a special attention (CBAM). Second, we leverage an adapted adversarial transformation of the learned representations that can generalize across different sketch domains (Table \ref{tab:domains}). This is enabled by the proposed losses.

\subsection{\textbf{Sketch-to-Image Synthesis Architecture}}\label{subsec:sketchtoimagearchitecture}
 
\begin{figure*}[http]
\centering
\includegraphics[width=1\textwidth]{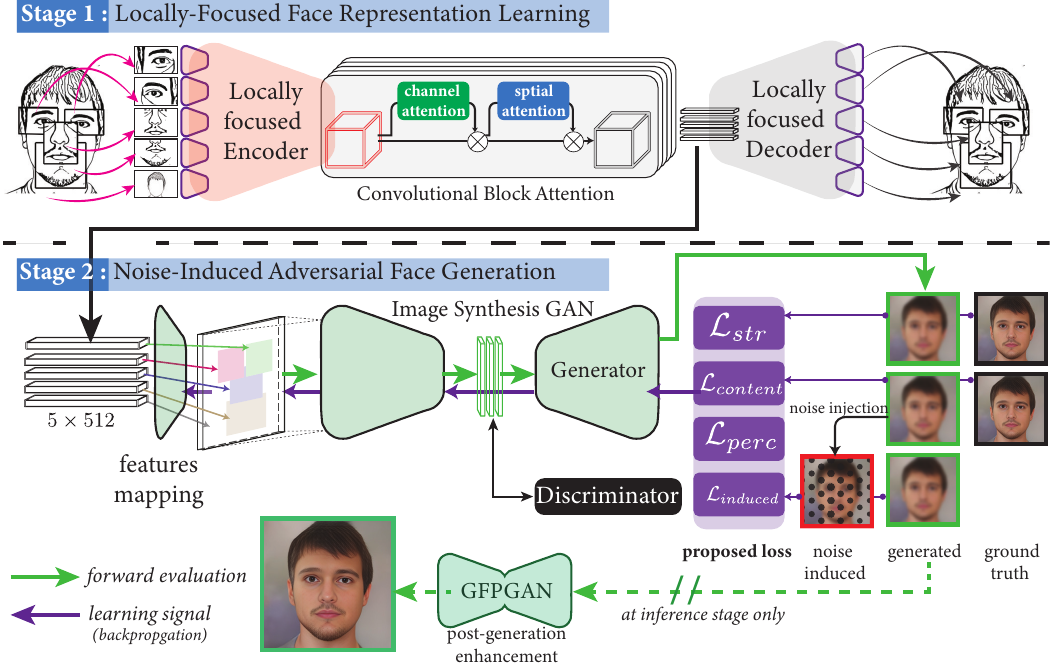}
\caption{Illustration 
of the proposed sketch to image generation architecture. (Top) \textbf{Locally-Focused Face Representation Learning}. The CBAM applies channel attention followed by spatial attention on each component of the sketch face to refine feature representations in the autoencoder. (Bottom) \textbf{Noise-Induced Adversarial Face Generation} . The feature descriptors of each facial component get converted into feature maps and these feature maps undergo the training process using enhanced cGAN. Finally, the generated image passed through a pre-trained GFPGAN ~\cite{wang2021towards} to enhance the quality of the generated image.}
\label{fig:architecture}
\end{figure*}

As shown in Fig.~\ref{fig:architecture},  we base the first stage on Auto-Encoder locally focused architecture integrating a convolution-based attention ~\cite{woo2018cbam}. This component is trained in self-supervised manner with the objective of reconstructing the input sketch.

The second stage maps the learned facial component features (feature mapping in Fig.~\ref{fig:architecture} into an input GAN model to generate the final face image to generate based on the face structure.

A global loss function is incorporated into the training process to preserve the shape of the generated image corresponding to the input sketch,
Finally, the generated image is fed into a pre-trained GFPGAN~\cite{wang2021towards} to enhance the quality of the generated face image. This step takes place only during the inference stage.

\subsubsection{\textbf{Locally-Focused Face Representation Learning }}

In this stage, we utilized component sketches $S^c$ to train five auto-encoders, each for generating unique feature embeddings against each component sketch as shown in \ref{fig:architecture}. Given the complex nature of the facial structure, we employ a patch-based strategy \cite{lepcha2023image} to divide the face into five components. Let $I_s$ and $I_p$ denote the sketch image and its color photo respectively.
\begin{equation*}
S^{c} = \textit{f}(I_{s})
\end{equation*}
Where \textit{f} is a static function that divides the face into five parts; $S^{c}, c \in \{1, 2, 3, 4, 5\}$, ``right eye'', ``left eye'', ``nose'', ``mouth'', and the ``remaining face''; respectively. 
After that, we learn a local feature embedding better to control the details of individual components of the face. We obtain the latent descriptors of individual components using five auto-encoder networks, denoted as $\{E_c, D_c\}$ with $E_c$ being an encoder and $D_c$ a decoder for component $c$. Each auto-encoder consists of five encoding layers and five decoding layers. We add a fully connected layer in the middle to ensure the latent descriptor is of 512 dimensions for all five components. 

Moreover, the model should generate a distinct image against each different sketch. For this purpose, we have introduced an additional Convolutional Block Attention Module (CBAM)~\cite{woo2018cbam} in the encoder architecture. The CBAM is an attention mechanism for convolutional neural networks (CNNs) that enhances performance by sequentially applying channel and spatial attention. It highlights important features in the input data, improving the network's ability to focus on relevant information.

\subsubsection{\textbf{Noise Induced Adversarial Face Generation }}
In the second training stage, the parameters of the trained component encoders are jointly fine-tuned within both the feature mapping and an adapted conditional GAN (cGAN) modules as shown in \ref{fig:architecture}.

In the features mapping (FM)  module, the five component encoders for different face components have distinct semantic meanings. The FM module is designed with five separate decoding models to convert feature vectors into spatial feature maps. Each decoding model includes a fully connected layer and five decoding layers. Each feature map contains 32 channels and matches the spatial size of the corresponding component in the sketch domain. The feature maps for the ``left-eye'', ``right-eye'', ``nose'', and ``mouth'' are placed back into the ``remainder'' feature maps according to the exact positions of the face components in the input face sketch image, preserving the original spatial relationships between the face components.

In the enhanced cGAN module, the combined feature maps are given to the GAN which converts them to realistic images. The $\Lb_{\text{content}}$ loss guides the generator to ensure the of quality pixel-wise output. Specifically, the loss is defined as :
\begin{equation}
\Lb_{\text{content}} = \frac{1}{N} \sum_{i=1}^{N} \left| G_i - R_i \right|
\end{equation}
where $ G_i$ and $R_i$ denotes the generated and the ground truth images respectively for $i \in [1, \cdots,  N]$ where $N$ is the number of samples in dataset.
 
In our proposed approach, we utilize several loss functions to guide the training of our neural network model.

For the conditional GAN loss, we use the standard adversarial loss:
\begin{equation}
\Lb_{\text{GAN}} = 
\mathbb{E}_{y \sim R} [\log d(y)] +\\ 
\mathbb{E}_{x \sim G} [\log (1 - d(g(x)))]
\end{equation}
where $g$ and $d$ denote the generator and discriminator respectively. Note that the GAN typically includes a discriminator loss. In our case, we use  Binary Cross Entropy (BCE) as the discriminator loss when classifying real vs fake. To ensure, semantic alignment (i.e. agreement at the feature space) between the generated and ground truth images, we further integrate a perceptual loss:

\begin{equation}
\text{$\Lb_{\text{perc}}$} = \frac{1}{N} \sum_{i=1}^{N} \left\| F(G_i) - F(R_i) \right\|
\end{equation}
Here, $F(.)$ represents a pre-trained VGG11 used to compute high-level features, $R_i$ and $G_i$ denote the original and generated image feature maps, respectively. After experimentation, we observed that by combining these loss functions, our framework is effectively guided during the training process and enables the generation of high-quality images from input sketches.

Yet, the system can perform poorly on unseen sketches from different domains (e.g. Line Sketch, Photoshop Sketch). To mitigate the issue, we propose discouraging the model from overfitting to the seen training domain. Specifically, we inject a noise on the generated image (from a uniform distribution). The idea is similar in spirit to adversarial perturbation \cite{poursaeed2018generative} yet with a different goal.

\begin{equation}
    I_{G^\prime} = I_{G} + \epsilon \times \mathbf{N} 
\end{equation}
\begin{equation}
    \Lb_{induced} = \frac{1}{N} \sum_{i=1}^{N} \left| I_{G^\prime} - I_{G} \right|
\end{equation}

where $I_{G} $ represents the image generated by the generator, $\epsilon$ is a small scaling factor, and $\mathbf{N}$ denotes a random noise sampled from a uniform distribution. Subsequently, we calculate the loss between the generated image and the perturbed image in which the noise is added. This guides the generator to enhance the overall visual quality without memorizing the noisy details of the images. Thus, improving generalization.

Finally, a structural loss is included to compute the luminance, contrast, and structural divergences between the generated images and input sketches.  Structural Similarity Index (SSIM) is used for this loss:
\begin{equation}
\Lb_{\text{str}}(R, G) = \frac{{(2\mu_R \mu_G + c_1)(2\sigma_{RG} + c_2)}}{{(\mu_R^2 + \mu_G^2 + c_1)(\sigma_R^2 + \sigma_G^2 + c_2)}}
\end{equation}
where $\mu_R$ and $\mu_G$ are the average pixel intensities of the real and generated images respectively, the variances of pixel intensities of real images and the generated images are denoted as $\sigma_R^2$ and $\sigma_G^2$ respectively, $\sigma_{RG}$ is the covariance of pixel intensities of real and the generated images and $c_1$ and $c_2$ are the small constants added to stabilize the division with weak denominator. The denominator terms are the local variances of the images, while the numerator terms are products of local means and covariances.

During the inference stage, the generated images from our trained network pass through pre-trained GFPGAN~\cite{wang2021towards} resulting in a substantial enhancement in image clarity, sharpness, and overall visual fidelity.

\begin{table}[h!]
\centering
\caption{Comparison of the quantitative results of our approach with other baseline techniques across three different datasets. The best scores are highlighted in bold black, while the next best scores are highlighted in bold blue.}
\label{tab:results}
\begin{tabular}{c@{\hspace{0.18cm}}c@{\hspace{0.18cm}}c@{\hspace{0.18cm}}c@{\hspace{0.18cm}}c@{\hspace{0.18cm}}c@{\hspace{0.18cm}}c@{\hspace{0.18cm}}c} 

 \hline
 \textbf{Method} & \textbf{FID} $\downarrow$ & \textbf{IS} $\uparrow$  &  \textbf{KID} $\downarrow$ & \textbf{SSIM} $\uparrow$ & \textbf{PSNR} $\uparrow$  \\ [0.5ex]
 \hline
 \hline
 \multicolumn{6}{c}{\textbf{CelebAMask-HQ dataset}} \\
   \hline
 PIx2PIxHD & 172.9270 & 1.2812    & 92.9465  & 0.6149 &  25.3097 \\
 CycleGAN & 148.7826 & 1.3118    & 105.989  & 0.5290 & 28.0248  \\
 pSp  & 99.15787 & \textbf{\textcolor{blue}{1.7411}}    & 73.5769  & \textbf{\textcolor{blue}{0.6582}} & \textbf{\textcolor{blue}{29.3249}}  \\
 OME  & - & - & - &  0.6171 & 21.4278 \\
 DeepFaceDrawing & \textbf{\textcolor{blue}{78.2768}} & 1.2512    & \textbf{\textcolor{blue}{61.2773}}  & 0.6401 & 28.9247 \\
  \textbf{Our}      & \textbf{61.8195}    & \textbf{1.7821}  & \textbf{60.6852} & \textbf{0.7984} & \textbf{30.4258} \\

 \hline
 \multicolumn{6}{c}{\textbf{CUHK dataset}} \\
 \hline
 PIx2PIxHD & 157.7083 & 1.3397    & 72.4963  & 0.6785 &  28.8489 \\

CycleGAN  & 176.6966 & 1.2244    & 69.3487 & 0.7616 & \textbf{28.9644} \\

pSp   & \textbf{\textcolor{blue}{129.4057}} &  \textbf{\textcolor{blue}{1.3700}}   & \textbf{\textcolor{blue}{34.1170}}  & \textbf{0.7847} & \textbf{\textcolor{blue}{28.6751}} \\
\textbf{Our}      & \textbf{106.4024}    & \textbf{1.4751} & \textbf{31.1037} & \textbf{\textcolor{blue}{0.7829}} & 28.6517 \\
\hline
 \multicolumn{6}{c}{\textbf{CUFSF dataset}} \\
 \hline
 PIx2PIxHD & 172.9593 & \textbf{\textcolor{blue}{1.67941}}    & 128.0782  &  0.6631 &  28.0573 \\
CycleGAN & \textbf{\textcolor{blue}{148.7460}} &  1.4112   & \textbf{\textcolor{blue}{78.2384}}  & 0.6858 & 28.0297 \\
pSp & 159.6944 & 1.6723    & 96.6299 & \textbf{0.7702} & \textbf{29.2038} \\

\textbf{Our}      & \textbf{110.9088}    & \textbf{1.8620} & \textbf{61.9566} & \textbf{\textcolor{blue}{0.7510}} & \textbf{\textcolor{blue}{28.2927}} \\
 \hline

\end{tabular}
\end{table}

\section{\textbf{Experiments and Results}}\label{sec:results and discussion}
\subsection{\textbf{Experimental Setting}}
\subsubsection{\textbf{Datasets}}\label{subsec:datapreparation}
We conduct extensive experiments of sketch-to-image generation on three benchmark datasets including the CelebAMask-HQ~\cite{lee2020maskgan}, CUHK~\cite{wang2008face}, and CUFSF~\cite{zhang2011coupled}. Moreover, we focus on generating a front face based on a sketch without ornamental items such as face masks or spectacles. Therefore, data preprocessing was performed to remove occluded faces. The goal is to convert face sketches into realistic face images, but the CelebAMask-HQ dataset includes full-body images of celebrities. To adapt these images for the model, face images were cropped using a pre-trained Haar cascade classifier~\cite{viola2001rapid}, commonly used for face recognition.

Furthermore, we require high-quality sketches without noise, so we applied a Photoshop photocopy filter along with sketch simplification technique \etal.~\cite{simo2016learning} to get sketches with reduced noise. After preprocessing, We have 11K, 667, and 188 sketch-image pairs for the CelebAMask-HQ, CUFSF, and CUHK datasets, respectively. Due to the relatively small size of our dataset, we aim to retain as much data as possible during training. Consequently, we adopted a 10:1 training-to-testing ratio in our studies.

\subsubsection{\textbf{Evaluation Metrics}}\label{subsubsec:evaluation metrics}

To evaluate the quality of our generated images, we employed heuristic-based evaluation measures commonly used in sketch-to-image translation tasks, including Inception Score (IS), Fréchet Inception Distance (FID), Kernel Inception Distance (KID), Peak Signal-to-Noise Ratio (PSNR) and Structural Similarity Index Measure (SSIM).

\begin{figure*}[http]
\centering

\rotatebox[origin=c]{90}{\parbox[c]{0\textheight}{\centering Input}}
\begin{minipage}[b]{0.14\textwidth}
  \centering
  \fbox{\includegraphics[width=\textwidth]{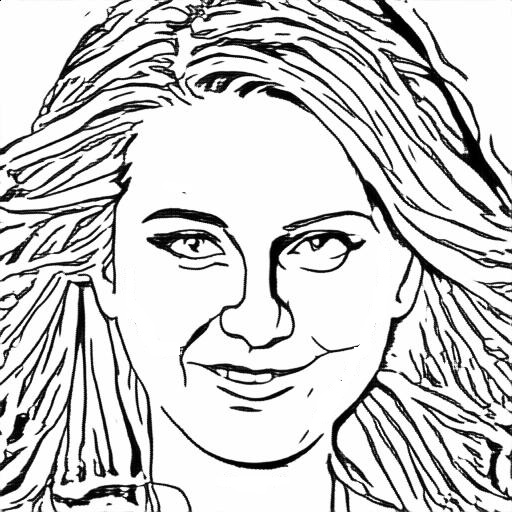}}
\end{minipage}%
\hspace{0.01\textwidth} 
\begin{minipage}[b]{0.14\textwidth}
  \centering
  \fbox{\includegraphics[width=\textwidth]{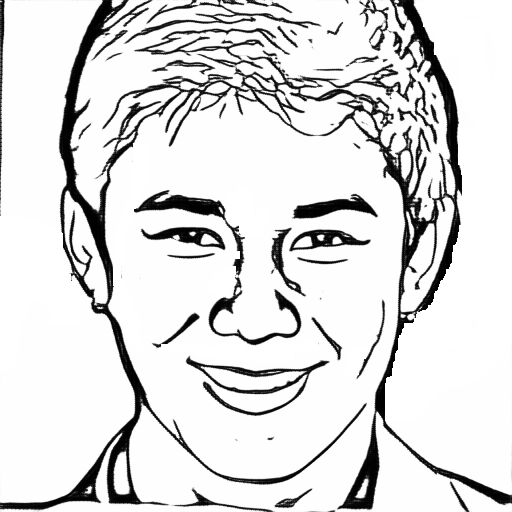}}
\end{minipage}%
\hspace{0.01\textwidth} 
\begin{minipage}[b]{0.14\textwidth}
  \centering
  \fbox{\includegraphics[width=\textwidth]{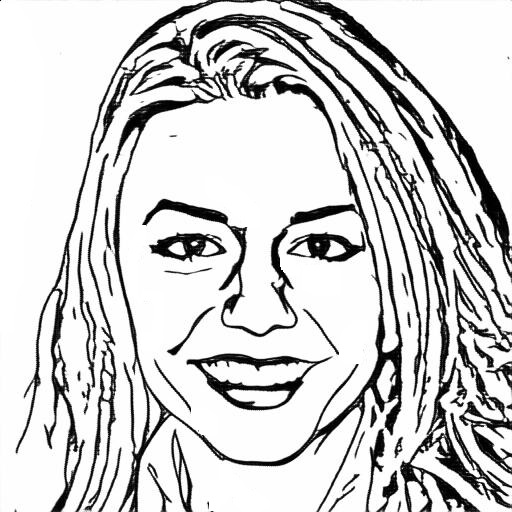}}
\end{minipage}%
\hspace{0.01\textwidth} 
\begin{minipage}[b]{0.14\textwidth}
  \centering
  \fbox{\includegraphics[width=\textwidth]{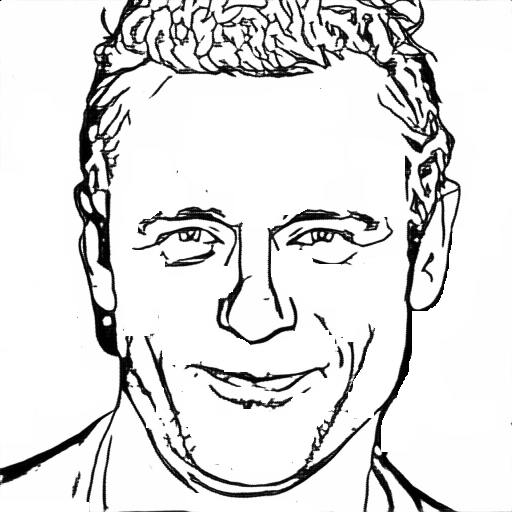}}
\end{minipage}%
\hspace{0.01\textwidth} 
\begin{minipage}[b]{0.14\textwidth}
  \centering
  \fbox{\includegraphics[width=\textwidth]{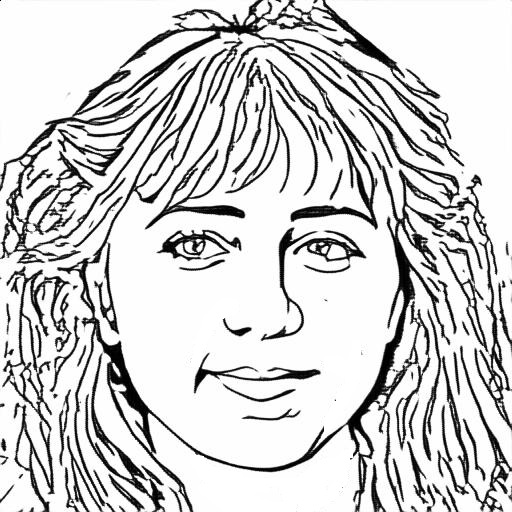}}
\end{minipage}%
\hspace{0.01\textwidth} 
\begin{minipage}[b]{0.14\textwidth}
  \centering
  \fbox{\includegraphics[width=\textwidth]{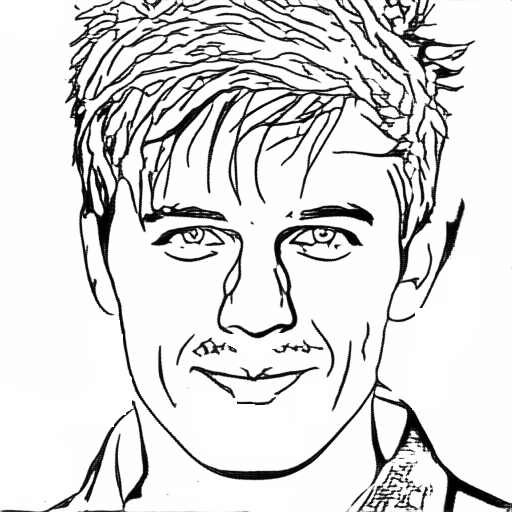}}
\end{minipage}
\hspace{0.05\textwidth} 

\vspace{0.5em} 

\rotatebox[origin=c]{90}{\parbox[c]{0\textheight}{\centering Label}}
\begin{minipage}[b]{0.14\textwidth}
  \centering
  \fbox{\includegraphics[width=\textwidth]{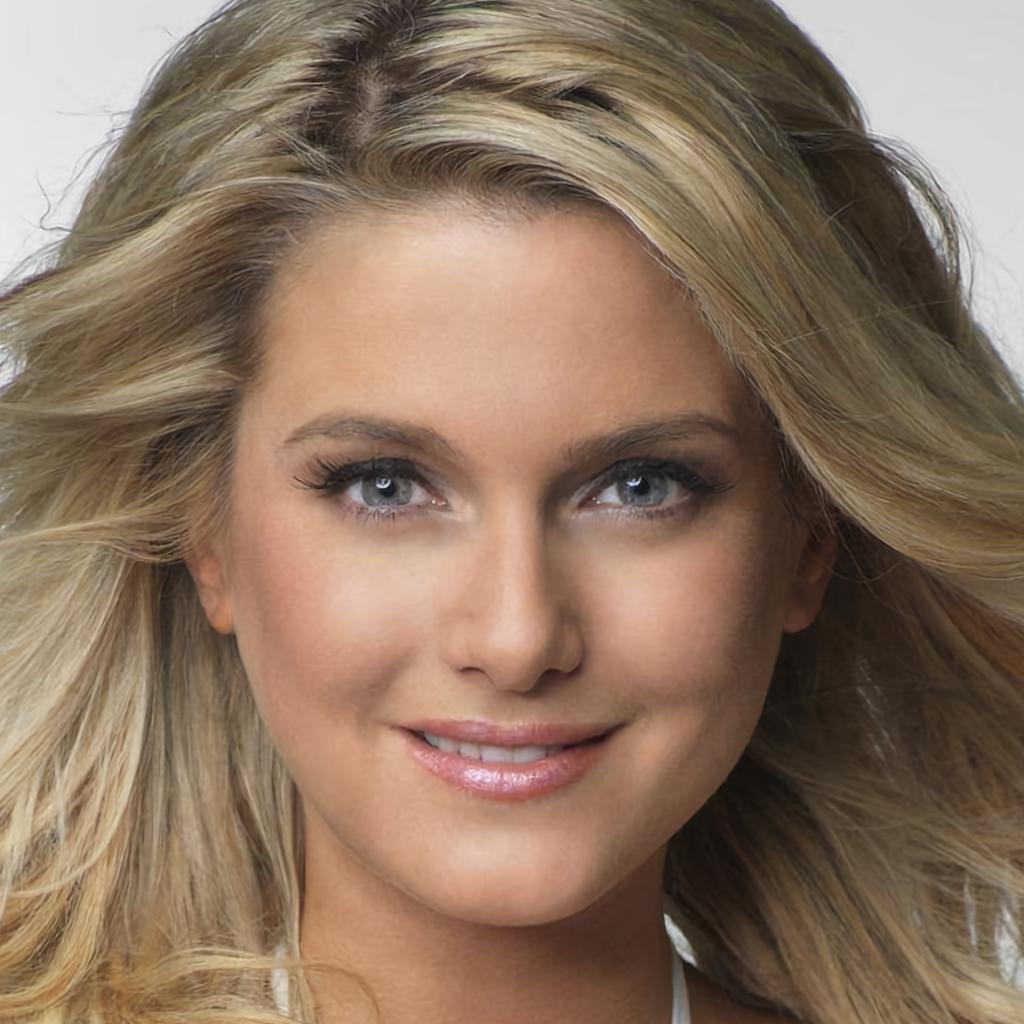}}
\end{minipage}%
\hspace{0.01\textwidth} 
\begin{minipage}[b]{0.14\textwidth}
  \centering
  \fbox{\includegraphics[width=\textwidth]{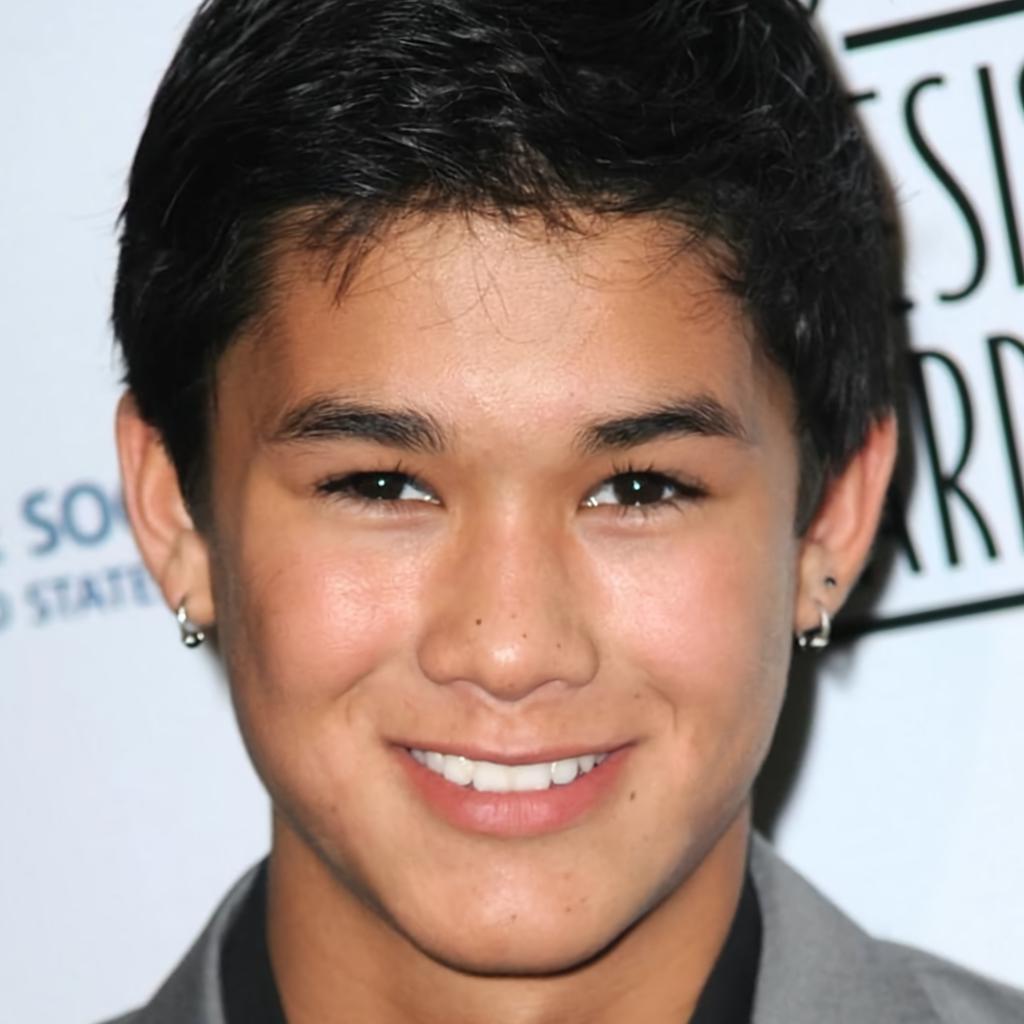}}
\end{minipage}%
\hspace{0.01\textwidth} 
\begin{minipage}[b]{0.14\textwidth}
  \centering
  \fbox{\includegraphics[width=\textwidth]{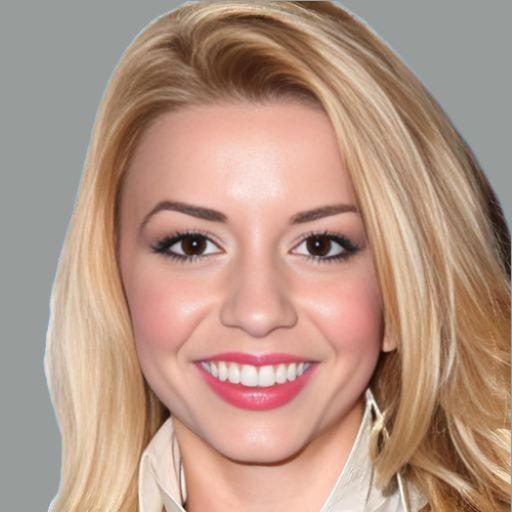}}
\end{minipage}%
\hspace{0.01\textwidth} 
\begin{minipage}[b]{0.14\textwidth}
  \centering
  \fbox{\includegraphics[width=\textwidth]{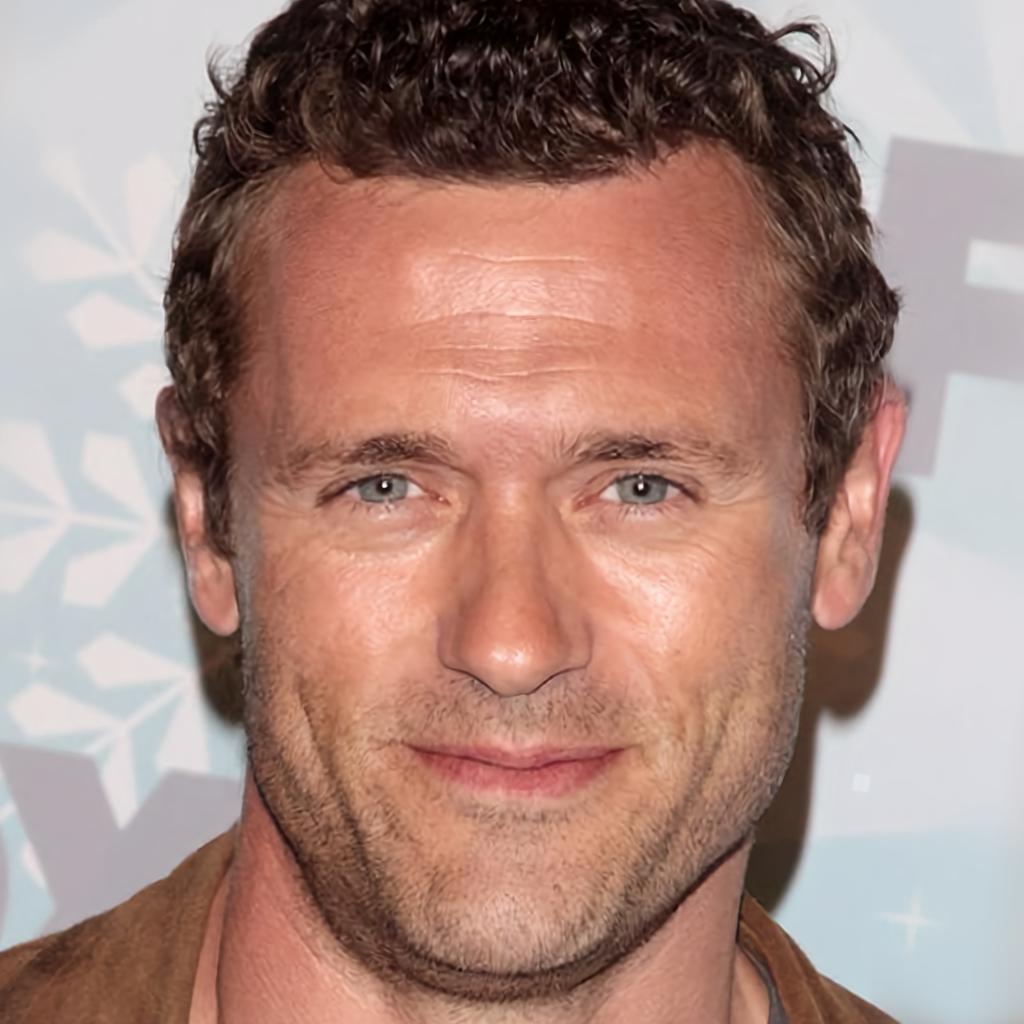}}
\end{minipage}%
\hspace{0.01\textwidth} 
\begin{minipage}[b]{0.14\textwidth}
  \centering
  \fbox{\includegraphics[width=\textwidth]{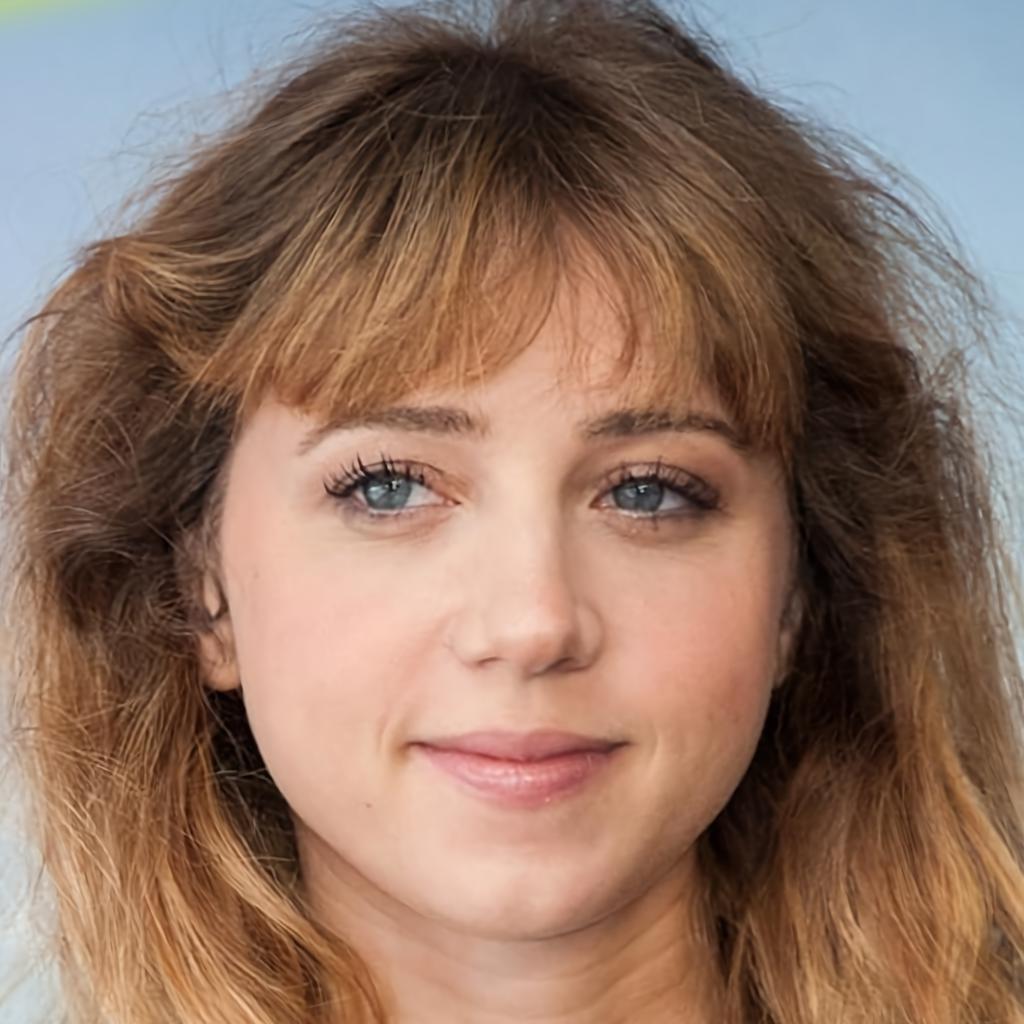}}
\end{minipage}%
\hspace{0.01\textwidth} 
\begin{minipage}[b]{0.14\textwidth}
  \centering
  \fbox{\includegraphics[width=\textwidth]{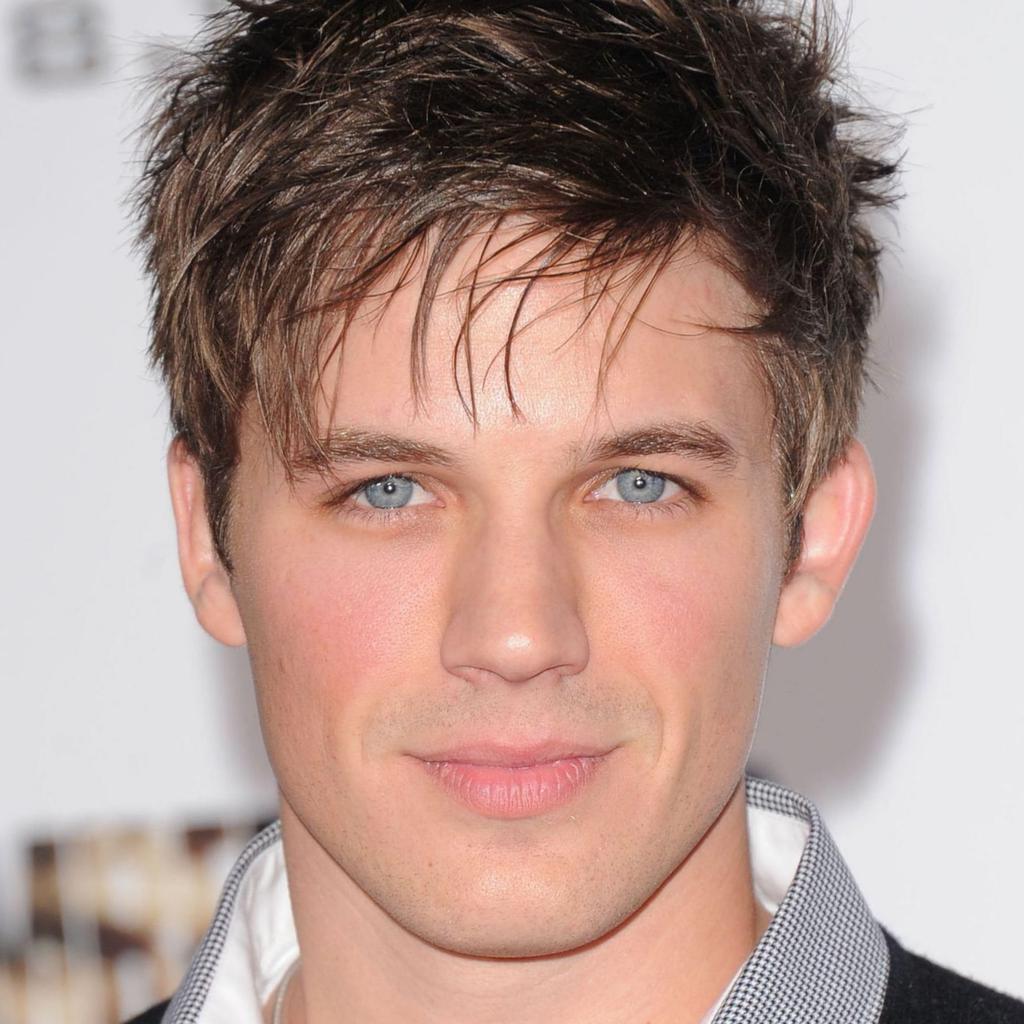}}
\end{minipage}
\hspace{0.05\textwidth} 

\vspace{0.5em} 

\rotatebox[origin=c]{90}{\parbox[c]{0\textheight}{\centering Pix2PixHD}}
\begin{minipage}[b]{0.14\textwidth}
  \centering
  \fbox{\includegraphics[width=\textwidth]{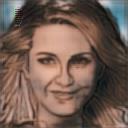}}
\end{minipage}%
\hspace{0.01\textwidth} 
\begin{minipage}[b]{0.14\textwidth}
  \centering
  \fbox{\includegraphics[width=\textwidth]{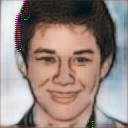}}
\end{minipage}%
\hspace{0.01\textwidth} 
\begin{minipage}[b]{0.14\textwidth}
  \centering
  \fbox{\includegraphics[width=\textwidth]{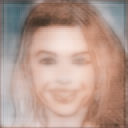}}
\end{minipage}%
\hspace{0.01\textwidth} 
\begin{minipage}[b]{0.14\textwidth}
  \centering
  \fbox{\includegraphics[width=\textwidth]{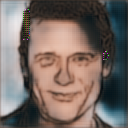}}
\end{minipage}%
\hspace{0.01\textwidth} 
\begin{minipage}[b]{0.14\textwidth}
  \centering
  \fbox{\includegraphics[width=\textwidth]{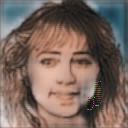}}
\end{minipage}%
\hspace{0.01\textwidth} 
\begin{minipage}[b]{0.14\textwidth}
  \centering
  \fbox{\includegraphics[width=\textwidth]{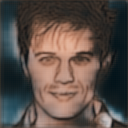}}
\end{minipage}
\hspace{0.05\textwidth} 

\vspace{0.5em} 
\rotatebox[origin=c]{90}{\parbox[c]{0\textheight}{\centering CycleGAN}}
\begin{minipage}[b]{0.14\textwidth}
  \centering
  \fbox{\includegraphics[width=\textwidth]{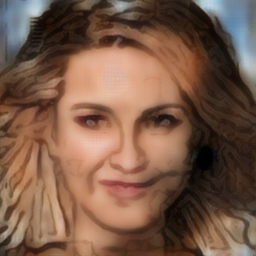}}
\end{minipage}%
\hspace{0.01\textwidth} 
\begin{minipage}[b]{0.14\textwidth}
  \centering
  \fbox{\includegraphics[width=\textwidth]{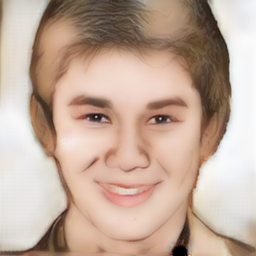}}
\end{minipage}%
\hspace{0.01\textwidth} 
\begin{minipage}[b]{0.14\textwidth}
  \centering
  \fbox{\includegraphics[width=\textwidth]{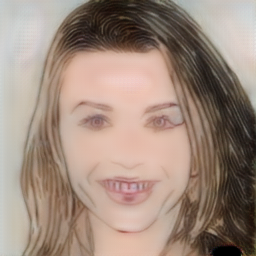}}
\end{minipage}%
\hspace{0.01\textwidth} 
\begin{minipage}[b]{0.14\textwidth}
  \centering
  \fbox{\includegraphics[width=\textwidth]{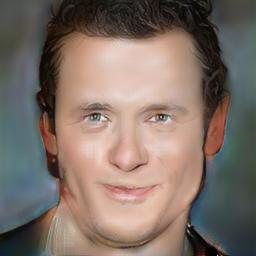}}
\end{minipage}%
\hspace{0.01\textwidth} 
\begin{minipage}[b]{0.14\textwidth}
  \centering
  \fbox{\includegraphics[width=\textwidth]{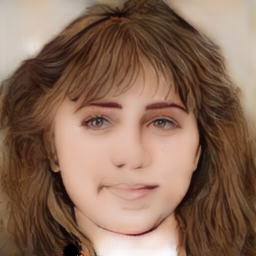}}
\end{minipage}%
\hspace{0.01\textwidth} 
\begin{minipage}[b]{0.14\textwidth}
  \centering
  \fbox{\includegraphics[width=\textwidth]{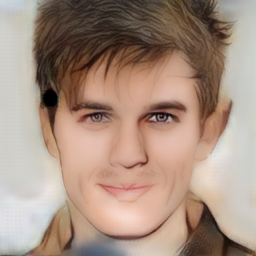}}
\end{minipage}
\hspace{0.05\textwidth} 

\vspace{0.5em} 
\rotatebox[origin=c]{90}{\parbox[c]{0\textheight}{\centering PSP}}
\begin{minipage}[b]{0.14\textwidth}
  \centering
  \fbox{\includegraphics[width=\textwidth]{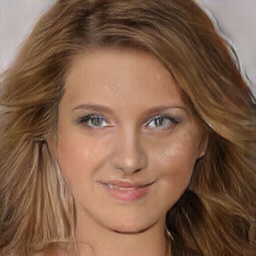}}
\end{minipage}%
\hspace{0.01\textwidth} 
\begin{minipage}[b]{0.14\textwidth}
  \centering
  \fbox{\includegraphics[width=\textwidth]{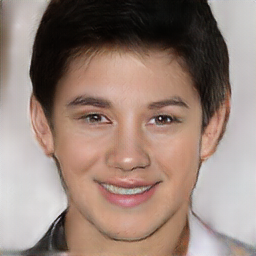}}
\end{minipage}%
\hspace{0.01\textwidth} 
\begin{minipage}[b]{0.14\textwidth}
  \centering
  \fbox{\includegraphics[width=\textwidth]{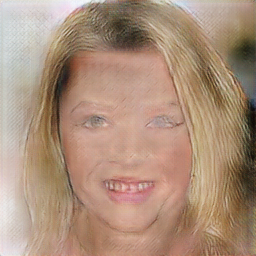}}
\end{minipage}%
\hspace{0.01\textwidth} 
\begin{minipage}[b]{0.14\textwidth}
  \centering
  \fbox{\includegraphics[width=\textwidth]{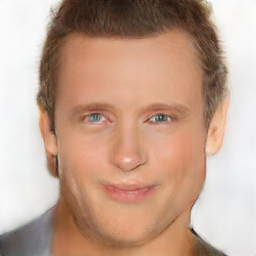}}
\end{minipage}%
\hspace{0.01\textwidth} 
\begin{minipage}[b]{0.14\textwidth}
  \centering
  \fbox{\includegraphics[width=\textwidth]{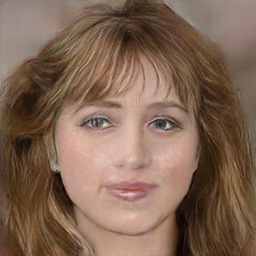}}
\end{minipage}%
\hspace{0.01\textwidth} 
\begin{minipage}[b]{0.14\textwidth}
  \centering
  \fbox{\includegraphics[width=\textwidth]{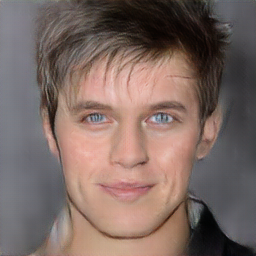}}
\end{minipage}
\hspace{0.05\textwidth} 

\vspace{0.5em} 
\rotatebox[origin=c]{90}{\parbox[c]{1cm}{\centering DeepFaceDrawing}}
\begin{minipage}[b]{0.14\textwidth}
  \centering
  \fbox{\includegraphics[width=\textwidth]{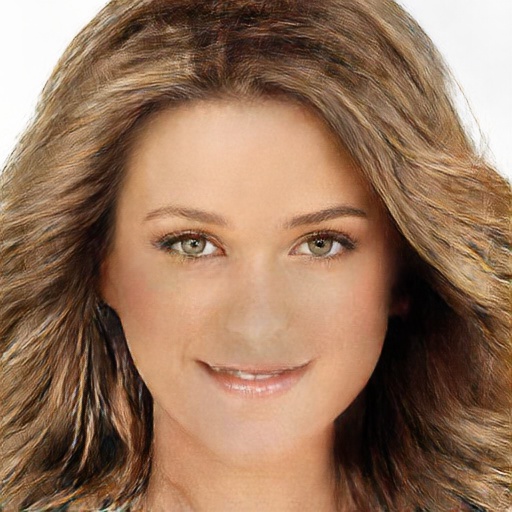}}
\end{minipage}%
\hspace{0.01\textwidth} 
\begin{minipage}[b]{0.14\textwidth}
  \centering
  \fbox{\includegraphics[width=\textwidth]{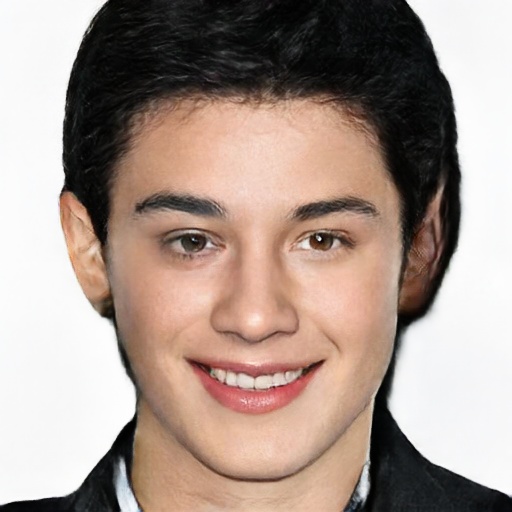}}
\end{minipage}%
\hspace{0.01\textwidth} 
\begin{minipage}[b]{0.14\textwidth}
  \centering
  \fbox{\includegraphics[width=\textwidth]{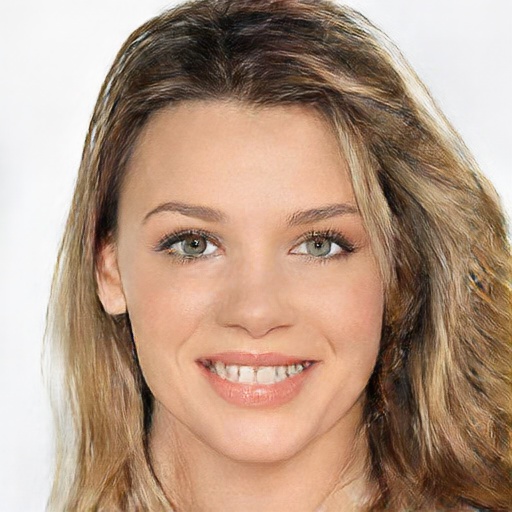}}
\end{minipage}%
\hspace{0.01\textwidth} 
\begin{minipage}[b]{0.14\textwidth}
  \centering
  \fbox{\includegraphics[width=\textwidth]{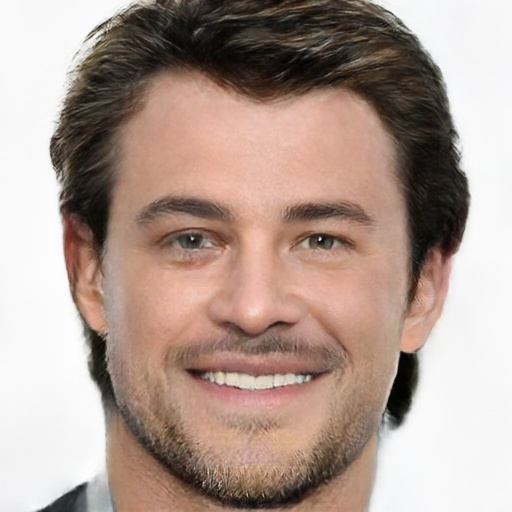}}
\end{minipage}%
\hspace{0.01\textwidth} 
\begin{minipage}[b]{0.14\textwidth}
  \centering
  \fbox{\includegraphics[width=\textwidth]{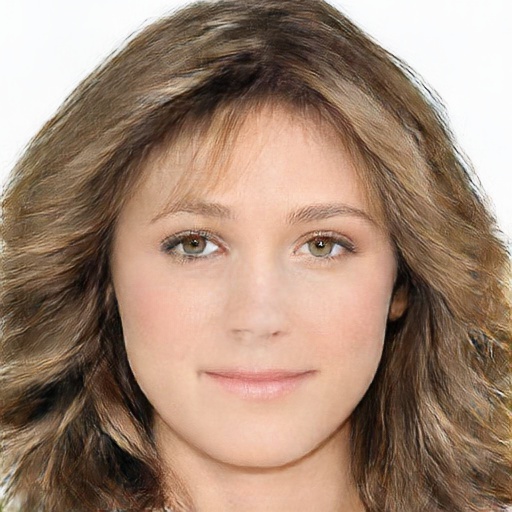}}
\end{minipage}%
\hspace{0.01\textwidth} 
\begin{minipage}[b]{0.14\textwidth}
  \centering
  \fbox{\includegraphics[width=\textwidth]{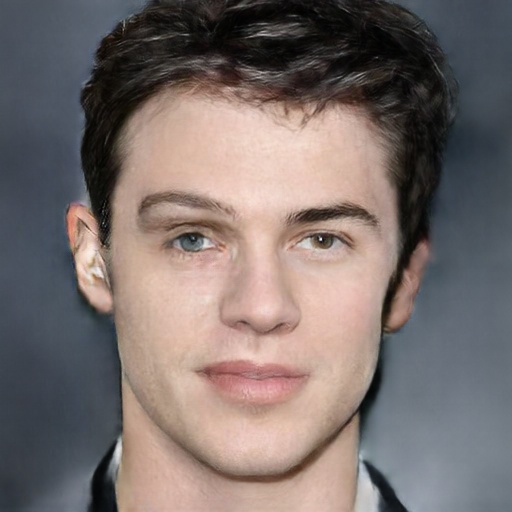}}
\end{minipage}
\hspace{0.05\textwidth} 

\vspace{0.5em} 
\rotatebox[origin=c]{90}{\parbox[c]{0\textheight}{\centering Ours}}
\begin{minipage}[b]{0.14\textwidth}
  \centering
  \fbox{\includegraphics[width=\textwidth]{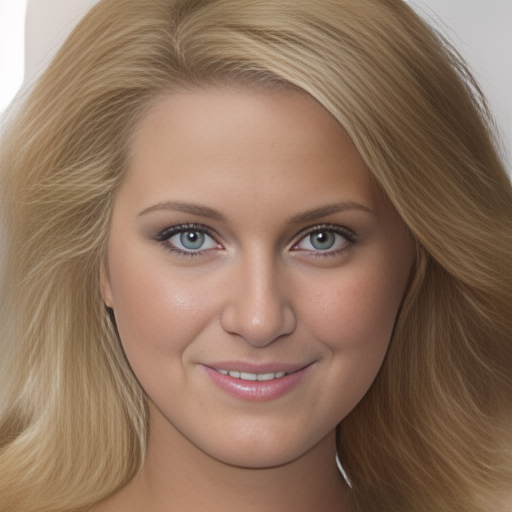}}
\end{minipage}%
\hspace{0.01\textwidth} 
\begin{minipage}[b]{0.14\textwidth}
  \centering
  \fbox{\includegraphics[width=\textwidth]{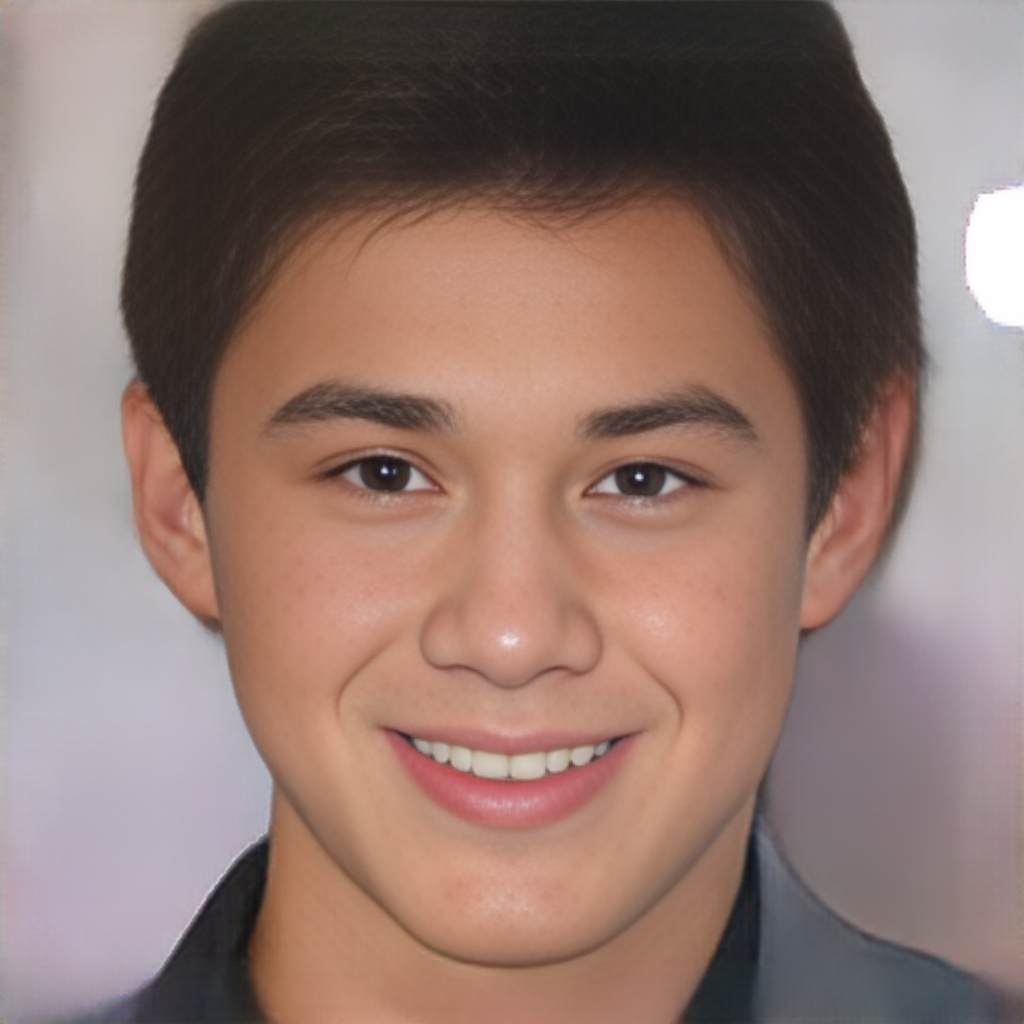}}
\end{minipage}%
\hspace{0.01\textwidth} 
\begin{minipage}[b]{0.14\textwidth}
  \centering
  \fbox{\includegraphics[width=\textwidth]{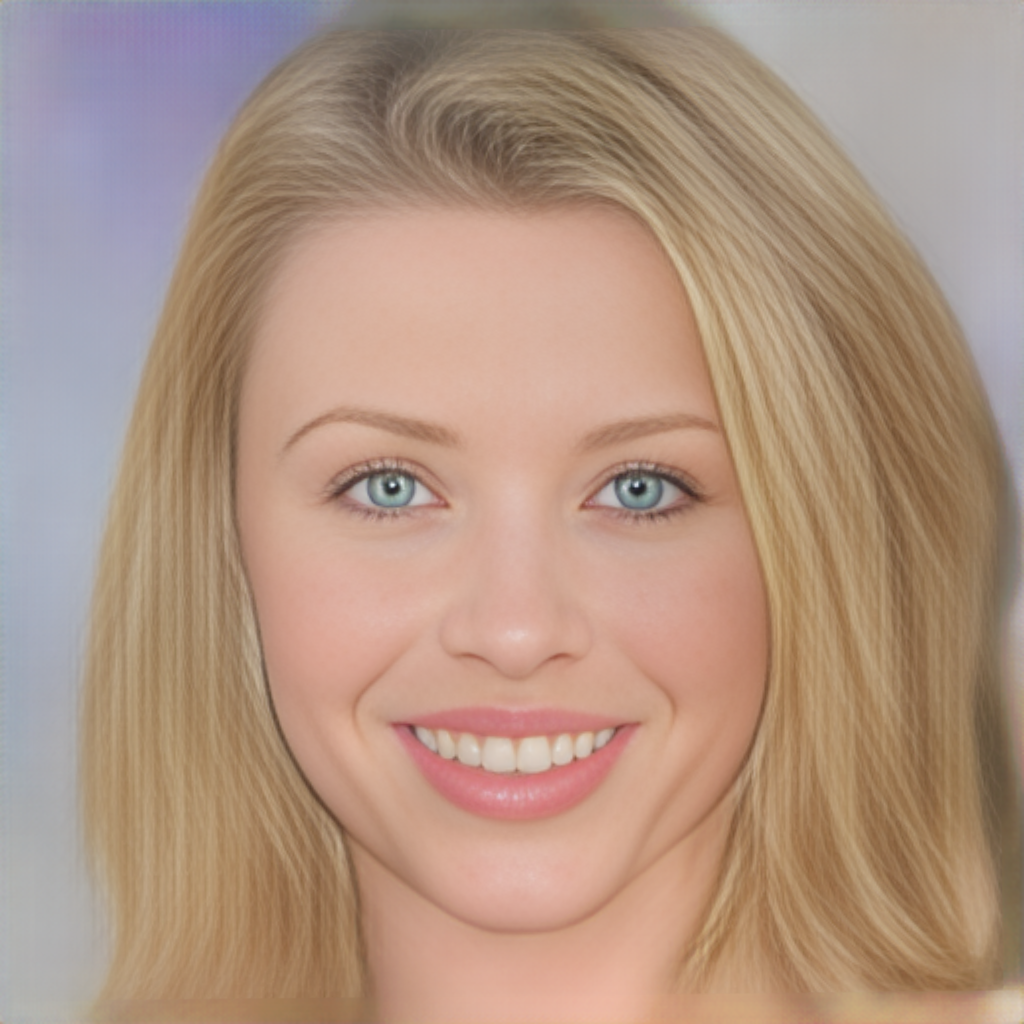}}
\end{minipage}%
\hspace{0.01\textwidth} 
\begin{minipage}[b]{0.14\textwidth}
  \centering
  \fbox{\includegraphics[width=\textwidth]{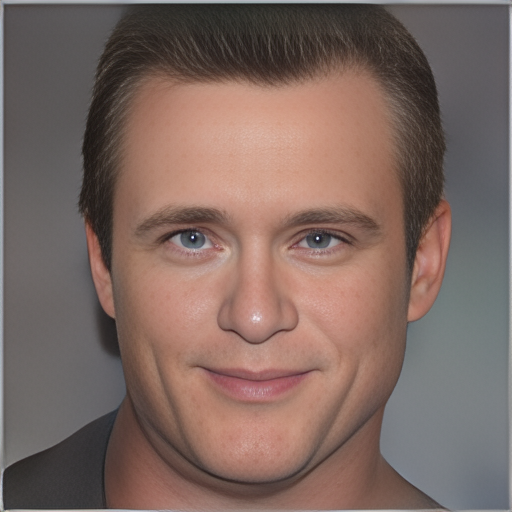}}
\end{minipage}%
\hspace{0.01\textwidth} 
\begin{minipage}[b]{0.14\textwidth}
  \centering
  \fbox{\includegraphics[width=\textwidth]{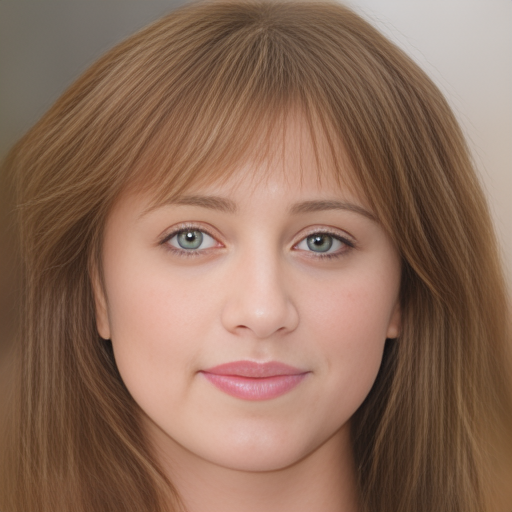}}
\end{minipage}%
\hspace{0.01\textwidth} 
\begin{minipage}[b]{0.14\textwidth}
  \centering
  \fbox{\includegraphics[width=\textwidth]{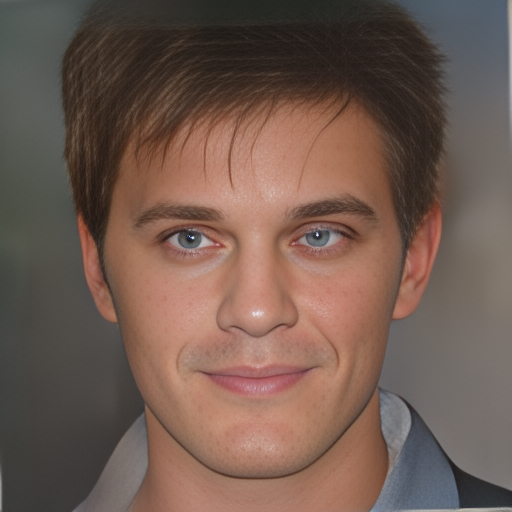}}
\end{minipage}
\hspace{0.05\textwidth} 

\caption{A qualitative comparison of results between our approach and other methods on CelebAMask-HQ dataset.}
\label{fig:qualitative results}
\end{figure*}
]

\begin{table}[h!]
\centering
\caption{Performance metrics of our framework under various ablation experiments.}
\label{tab:domains}
\label{table:ablation_quantitave}
\begin{tabular}{@{}cccc|ccccc@{}}
 \hline
 CBAM & DA & GL & IE & FID $\downarrow$ & IS $\uparrow$ & KID $\downarrow$ & SSIM $\uparrow$ & PSNR $\uparrow$ \\ 
 \hline\hline
  & & & & 115.84 & 1.53 & 66.75 & 0.67 & 18.02 \\
  & & & \ding{51} & 107.90 & 1.61 & 65.42 & 0.73 & 20.21 \\
  & \ding{51} & \ding{51} & & 60.77 & 1.61 & 62.67 & 0.78 & 23.02 \\
  & \ding{51} & \ding{51} & \ding{51} & \textbf{56.77} & \textbf{1.88} & 60.80 & 0.79 & 25.09 \\ 
 \ding{51} & & & & 102.20 & 1.67 & 61.49 & 0.74 & 17.31 \\
 \ding{51} & & & \ding{51} & 86.14 & 1.70 & 63.29 & 0.73 & 21.28 \\
 \ding{51} & \ding{51} & \ding{51} & & 74.89 & 1.74 & 60.74 & 0.79 & 29.69 \\
 \ding{51} & \ding{51} & \ding{51} & \ding{51} & 61.82 & 1.79 & \textbf{60.69} & \textbf{0.80} & \textbf{30.43} \\
 \hline
\end{tabular}
\end{table}

These metrics provide insights into various aspects of the generated images, such as diversity, fidelity, and structural similarity.

\subsection{\textbf{Quantitative Results}}\label{subsec:quantitative results}

We first provide a quantitative comparison of our framework against state-of-the-art techniques on three benchmark datasets as shown in Table \ref{tab:results}. These techniques includes PIx2PIxHD \cite{wang2018high}, CycleGAN \cite{zhu2017unpaired}, pSp \cite{richardson2021encoding}, OME \cite{guo2024image} and the DeepFaceDrawing \cite{chen2020deepfacedrawing}. Notably, our framework achieves superior performance across all evaluated metrics, with significantly lower FID, KID, and higher IS, SSIM, and PSNR scores. These results underscore the effectiveness of our approach in generating faces that closely resemble real faces in terms of both appearance and structural similarity.


On the CelebAMask-HQ dataset, our framework shows 21\%, 42\%, 0.84\%, 23\%, and 5\%  improvement in FID, IS, KID, SSIM, and PSNR scores against DeepFaceDrawing respectively. Moreover, there has been a significant improvement achieved by our method against other methods as shown in Table \ref{tab:results}. Additionally, we trained our framework on the CUHK and CUFSF datasets and compared its performance to techniques like pSp, CycleGAN, and PIx2PIxHD. Since the training codes for OME and DeepFaceDrawing methods are not publicly available, we only had access to their weights for the CelebAMask-HQ dataset. 
On the CUHK dataset, our framework showed improvements of 21\% in FID, 6.8\% in IS, and 9.6\% in KID scores compared to pSp. However, pSp performed slightly better in terms of SSIM, and CycleGAN performed slightly better in terms of PSNR. Moreover, on the CUFSF dataset,  there is an improvement of 43\% in FID, 11\% in IS, and 56\% in KID scores compared to the pSp method. However, pSp showed better results on the CUFSF dataset regarding SSIM and PSNR as shown in Table \ref{tab:results}.

\subsubsection{\textbf{Results on Different type of Sketches}}\label{subsubsec:different sketch results}
To highlight the generalization improvement in the performance of our framework, we present the results on three different types of sketches. These sketches vary in style, complexity, and source. The types include hand-drawn sketches, line sketches, and Photoshop sketches. The hand-drawn sketches are created by human artists, while line sketches and Photoshop sketches are computer-aided. 

\begin{table}[h!]
\centering
\caption{Results of our framework on different types of sketches show the domain invariance of our framework.}
\label{tab:domain_adaptation_results}
\begin{tabular}{|l|l|l|l|l|l|}
\hline
\textbf{Sketch Type} & \textbf{FID} $\downarrow$ & \textbf{IS} $\uparrow$ & \textbf{KID} $\downarrow$ & \textbf{SSIM} $\uparrow$ & \textbf{PSNR} $\uparrow$\\ \hline
\textbf{Hand Drawn Sketch}        & 135.57         & 1.61        & 61.31        & 0.68         & 27.88         \\ \hline
\textbf{Line Sketch}      & 83.86        & 1.55        & 43.38        & 0.75          & 27.90         \\ \hline
\textbf{Photoshop Sketch}      & 79.35        & 1.70        & 45.51        & 0.76          & 27.92         \\ \hline

\end{tabular}
\end{table}

\begin{table}[h!]
\centering
\caption{Results of DeepFaceDrawing on different types of sketches.}
\label{tab:domain_adaptation_results_DFD}
\begin{tabular}{|l|l|l|l|l|l|}
\hline
\textbf{Sketch Type} & \textbf{FID} $\downarrow$ & \textbf{IS} $\uparrow$ & \textbf{KID} $\downarrow$ & \textbf{SSIM} $\uparrow$ & \textbf{PSNR} $\uparrow$\\ \hline
\textbf{Hand Drawn Sketch}        & 191.01         & 1.70        & 94.65        & 0.52         & 25.98         \\ \hline
\textbf{Line Sketch}      & 130.88        & 1.49        & 38.87        & 0.67          & 25.47         \\ \hline
\textbf{Photoshop Sketch}      & 115.63        & 1.51        & 59.15        & 0.65          & 25.80         \\ \hline

\end{tabular}
\end{table}

Table \ref{tab:domain_adaptation_results} shows the results of our framework, trained on the CelebAMask-HQ dataset and tested on 188 sketches from the CUHK dataset of hand-drawn sketches. The other two types of sketch datasets were created using 188 images from the CUHK dataset. The results of DeepFaceDrawing on these types of sketches are also shown in Table \ref{tab:domain_adaptation_results_DFD}.

Our framework performed well across all three sketch categories against all evaluation measures. Especially in the Photoshop and line sketch categories, there is a 45\% and 56\% improvement in FID score compared with DeepFaceDrawing respectively. The reason is these sketches are computer-aided face sketches and resemble more with the original images. However, the hand-drawn sketches created by human artists are not perfect. These results demonstrate that our framework has generalization capability and can generate images based on different types of sketches.

\subsubsection{\textbf{Ablation Study}}\label{subsec:ablationexperiments}

We performed an ablation study on key modules, namely Convolutional Block Attention Module (CBAM), Domain Adaptation (DA), Global Loss (GL), and Image Enhancement (IE), To validate the efficacy of our framework using the CelebAMask-HQ dataset. As demonstrated in Table \ref{table:ablation_quantitave}, the absence of all modules degrades the performance of the framework.

The noise addition in the generated image and the computing loss between the generated image and the noisy image in the DA module combined with the global loss in the GL module during training give good results. In particular, including the IE module significantly boosts the overall performance of our network. In addition, the combination of DA, GL, and IE results in improved FID and IS scores. A comparative analysis underscores the consistent enhancement in the quality of the generated images achieved by our framework. 


\subsection{\textbf{Qualitative Results}}
Our qualitative analysis further supports the quantitative findings, as depicted in Figure \ref{fig:qualitative results}. The results of Pix2PixHD and CycleGAN effectively capture facial expressions. However, the generated images exhibit some dark patches. On the other hand, DeepFaceDrawing produces brighter images, but they do not accurately correspond to the ground truth labels as shown in 6$^{th}$ row and 4$^{th}$ column in Figure \ref{fig:qualitative results}. Upon examination, it becomes evident that the faces generated by our framework exhibit a higher degree of realism than those produced by DeepFaceDrawing and other techniques of sketch-to-image synthesis.


Importantly, our qualitative superiority is also validated by feedback from human evaluators. Through human-centric assessments, our generated faces consistently received higher ratings in terms of realism, expressiveness, and overall visual appeal, underscoring the effectiveness of our framework in creating socially and culturally relevant faces.


\section{\textbf{CONCLUSION and FUTURE WORK}}
\label{sec:Conclusion}

This work introduces a novel two-step approach that addresses the limitations inherent in traditional sketch-to-image processes. The framework is specifically designed to enhance locally focused face representation by independently identifying and processing five key facial feature descriptors. This approach ensures that crucial facial components are not only accurately represented but also precisely reconstructed. The integration of innovative loss functions and noise-induced refinement techniques significantly enhances the detail and realism of the generated images. Comparative analysis against five established methods demonstrates the superior performance of our framework, affirming its efficacy. Additionally, the framework exhibits robust performance across various sketch types, highlighting its generalizability and practical applicability, particularly for law enforcement agencies in criminal identification. As technology continues to advance, our approach underscores the substantial potential of deep learning-based models in addressing real-world challenges. Future enhancements in generalization performance across different sketch styles could be achieved through the incorporation of advanced domain generalization techniques or adversarial training methods.

\bibliographystyle{ieeetr} 
\bibliography{main}

\end{document}